\documentclass[journal]{IEEEtran}
\usepackage{cite}
\usepackage{graphicx}
\usepackage{amsmath}
\usepackage{flushend}
\usepackage{subfigure}
\usepackage{amssymb}
\usepackage{booktabs}
\usepackage{tikz}

\makeatletter
\let\NAT@parse\undefined
\makeatother
\usepackage{hyperref}  

\ifCLASSINFOpdf
\else
\fi
\usepackage{tikz,xcolor,hyperref}
\definecolor{lime}{HTML}{A6CE39}
\DeclareRobustCommand{\orcidicon}{%
    \begin{tikzpicture}
    \draw[lime, fill=lime] (0,0) 
    circle [radius=0.16] 
    node[white] {{\fontfamily{qag}\selectfont \tiny ID}};    \draw[white, fill=white] (-0.0625,0.095) 
    circle [radius=0.007];    \end{tikzpicture}
    \hspace{-2mm}}
\foreach \x in {A, ..., Z}{%
    \expandafter\xdef\csname orcid\x\endcsname{\noexpand\href{https://orcid.org/\csname orcidauthor\x\endcsname}{\noexpand\orcidicon}}
    }

\begin{document}
%
\title{Complementary Calibration: Boosting General Continual Learning with Collaborative  Distillation and Self-Supervision}
%
%
%
\newcommand{\orcidauthorA}{0000-0002-2197-3739}

\author{Zhong~Ji\orcidA{}, \textit{Senior Member, IEEE},
        Jin~Li,
        Qiang~Wang*,
        and~Zhongfei~Zhang, \textit{Fellow, IEEE}

\thanks{Manuscript received xxx xx, 2021; revised xxx xx, 2021.}
\thanks{This work was supported by the National Natural Science Foundation of China (NSFC) under Grants 6217020340 and 61771329.}
\thanks{Zhong Ji, Jin Li, and Qiang Wang are with the School of Electrical and Information Engineering, Tianjin University, Tianjin 300072, China, and also with the Tianjin Key Laboratory of Brain-Inspired Intelligence Technology, Tianjin University, Tianjin 300072, China (e-mail: jizhong@tju.edu.cn; lijincm@tju.edu.cn; qiangwang306@tju.edu.cn).}
\thanks{Zhongfei Zhang is with the Computer Science Department, Binghamton University, State University of New York, Binghamton, NY 13902, USA (e-mail: zzhang@binghamton.edu).}
\thanks{*The corresponding author is Qiang Wang.}
}

\markboth{Submitted to IEEE TRANSACTIONS ON IMAGE PROCESSING,~Vol.~XX, No.~X, August~20XX}%
{Ji \MakeLowercase{\textit{et al.}}: Complementary Calibration: Boosting General Continual Learning with Collaborative  Distillation and Self-Supervision}

\maketitle

\begin{abstract}
General Continual Learning (GCL) aims at learning from non independent and identically distributed stream data without catastrophic forgetting of the old tasks that don’t rely on task boundaries during both training and testing stages. We reveal that the relation and feature deviations are crucial problems for catastrophic forgetting, in which relation deviation refers to the deficiency of the relationship among all classes in knowledge distillation, and feature deviation refers to indiscriminative feature representations. To this end, we propose a Complementary Calibration (CoCa) framework by mining the complementary model’s outputs and features to alleviate the 
two deviations in the process of GCL. Specifically, we propose a new collaborative distillation approach for addressing the relation deviation. It distills model's outputs by utilizing ensemble dark knowledge of new model's outputs and reserved outputs, which maintains the performance of old tasks as well as balancing the relationship among all classes. Furthermore, we explore a  collaborative self-supervision idea to leverage pretext tasks and supervised contrastive learning for addressing the feature deviation problem by learning complete and discriminative features for all classes. Extensive experiments on four popular datasets show that our CoCa framework achieves superior performance against state-of-the-art methods. 
\end{abstract}


%
\IEEEpeerreviewmaketitle

\section{Introduction}
%
%
%
%
\IEEEPARstart{H}{uman}  beings have the capability to continuously acquire, adjust and transfer knowledge, which is just we desire agents to have. Continual learning  \cite{review0}, \cite{review_pami}, also called incremental learning or lifelong learning, focuses on the problem of learning from a data stream in non-stationary data distributions. 
These data come from different tasks, in which the input domains are constantly changing. In this situation, we hope to exploit the acquired knowledge to solve the old and new tasks. Continual learning has a wide range of related applications in the real world, such as object detection \cite{did}, product search \cite{MetaSearch} and 3D object classification \cite{L3DOC}. \par
The main challenge in continual learning is catastrophic forgetting \cite{cf1989}, that is, when a deep neural network is trained on a new task, the performance on old tasks usually drops significantly. To prevent the catastrophic forgetting, we hope that the model is capable to integrate new and existing knowledge from new data (plasticity) on the one hand, and prevent the significant interference of new input on existing knowledge (stability) on the other hand. These two conflicting demands constitute the stability-plasticity dilemma \cite{review0}.\par
\begin{figure}
	\begin{center}
		\includegraphics[scale=0.7,trim=25 10 10 20,clip]{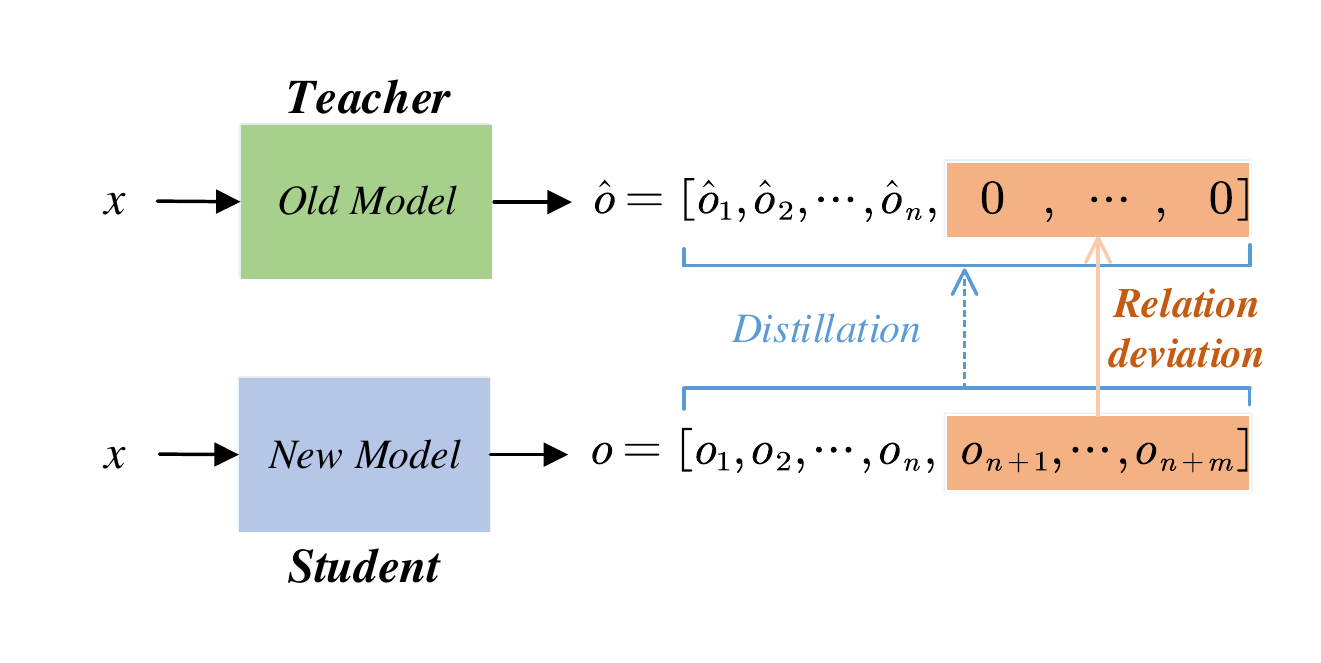}
	\end{center}
	\caption{Diagram of relation deviation in knowledge distillation. The old model has not seen new classes, thus its output $\hat{o}$ lacks of the relationship between old and new classes. When the reserved sample $x$ is replayed, the deficiency of relationship results in a deviation between the output of the new model $o$ and the output of the old model $\hat{o}$, which leads to inaccurate continual learning.}
	\label{fig:mot1}
\end{figure}
Early studies of continual learning primarily focused on the Task Incremental Learning (Task-IL) scenario \cite{GEM}, \cite{EWC}, \cite{SI}, in which the difficulty is greatly reduced by employing task boundaries during testing stage.  Recently, lots of studies consider a more challenging setting, i.e., Class Incremental Learning (Class-IL) \cite{lucir}, \cite{podnet}, \cite{dde}, in which task boundaries are unavailable when testing stage. However, existing methods for both Task-IL and Class-IL rely on task boundaries in the training stage, which are not in line with the practical requirement. In this paper, we consider a more complex and practical setting: General Continual Learning (GCL) \cite{review_pami}, \cite{NEURIPS2020_DER}, whose task boundaries are not available during both training and testing stages. Therefore, most of the existing continual learning methods  cannot be applied to GCL.\par
Recently, Buzzega \textit{et al.} \cite{NEURIPS2020_DER} proposed a simple and strong GCL baseline named Dark Experience Replay (DER). They balanced the stability-plasticity dilemma by knowledge distillation. Concretely, they took old model as the teacher and reserved the old model's outputs to constrain the new model's outputs of the old samples. However, new samples are unseen to the old model, which lead to inaccurate old model's outputs. As shown in Fig. \ref{fig:mot1}, the outputs of an old model lack the relationship between the old  and new classes. In addition, when a new task consists of  new samples of the old classes, the relationship among the classes in the old model may incomplete. We refer to the deficiency of the relationship among all classes in knowledge distillation as relation deviation, which makes interference in balancing the relationship of the old and new classes. \par
\begin{figure}
	\begin{center}
		\includegraphics[scale=0.55,trim=20 20 10 20, clip]{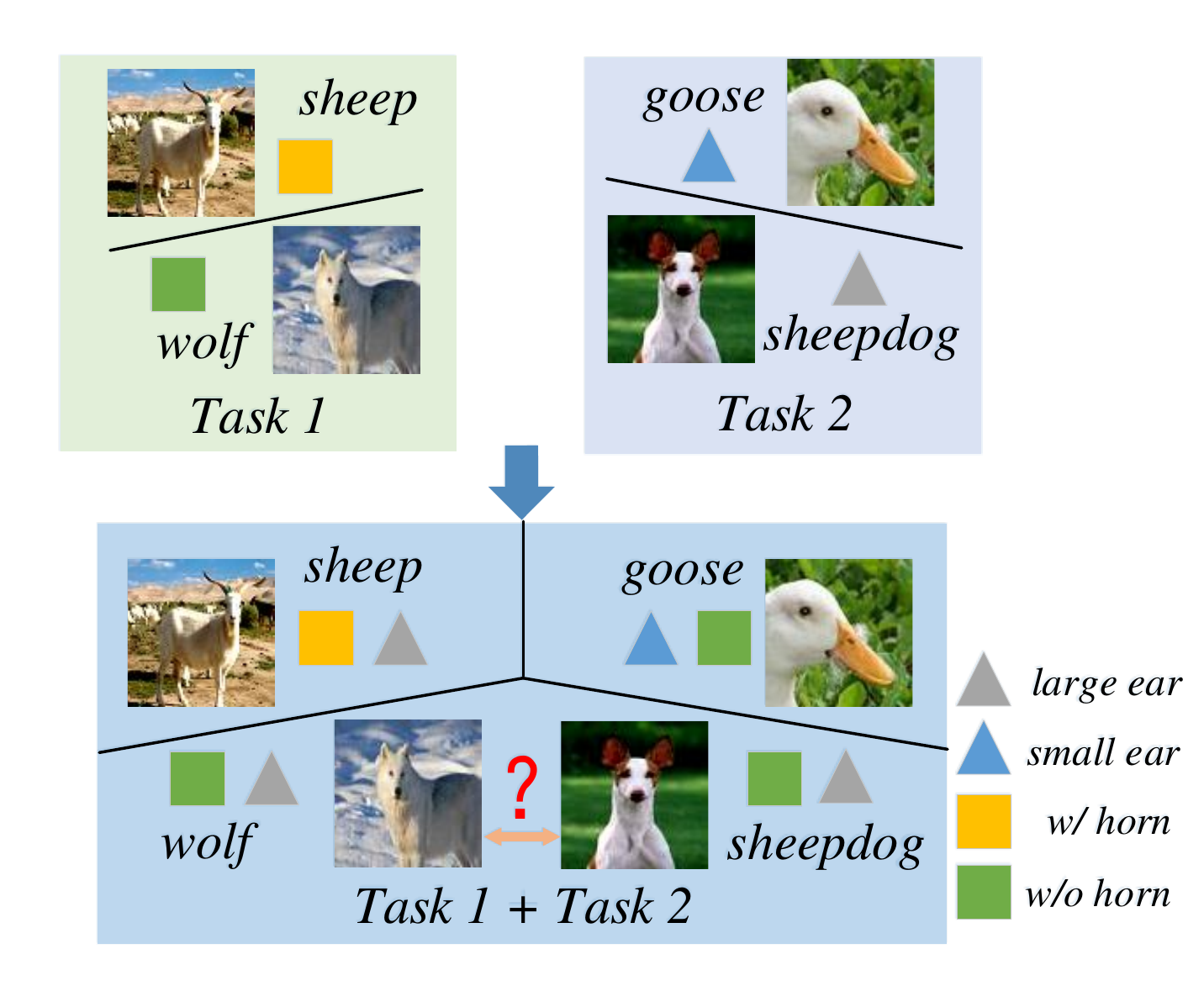}
	\end{center}
	\caption{Diagram of the indiscriminative feature representations in GCL. The  feature extractor tends to extract discriminative feature for a single task  rather than all tasks. For example, the "horn" feature is enough to distinguish sheep from wolf for task 1, while the "ear" feature is enough to distinguish the sheepdog and goose for task 2. However, when both tasks are mixed, previously extracted features may not be discriminative enough, that is, the feature representations are incomplete. For example, the "ear" and "horn" features are indiscriminative in representing sheepdog and wolf.}
	\label{fig:mot2}
\end{figure}
Moreover, another reason for catastrophic forgetting in GCL is that the feature representations are not discriminative enough in representing both the old and new classes, that is, the feature representations are incomplete for all classes in GCL. As shown in Fig. \ref{fig:mot2}, the  feature extractor tends to extract the discriminative features of a single task, which are indiscriminative in representing the data in the following classes. Thus, it is hard to distinguish all classes with them. We name this indiscriminative  feature representation problem as feature deviation.\par
\begin{figure*}
	\begin{center}
		\includegraphics[scale=0.66,trim=22 20 20 18,clip]{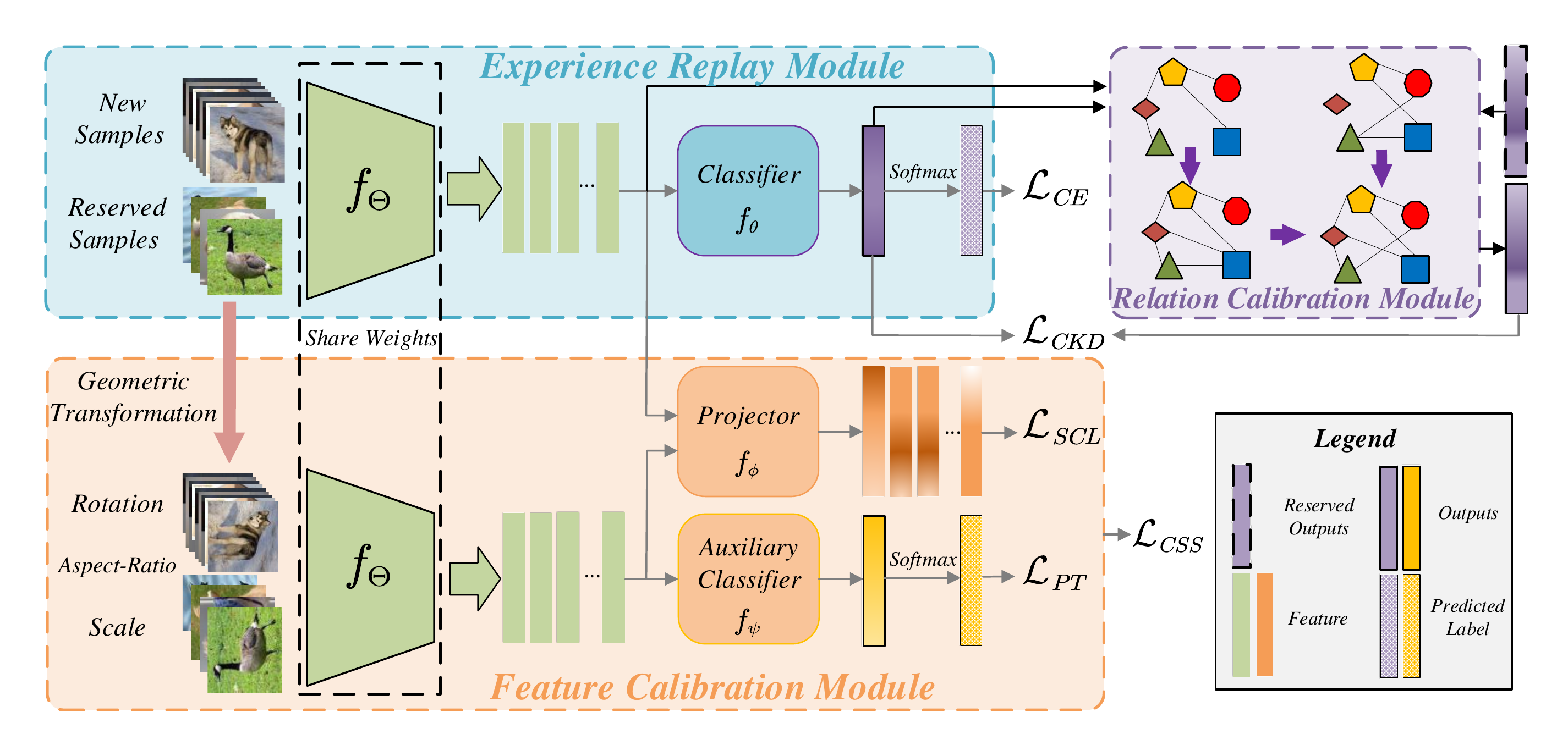}
	\end{center}
	\caption{Overview of the proposed CoCa framework at the training stage, which is composed of three parts: an Experience Replay Module, a Relation Calibration Module, and a Feature Calibration Module. A backbone $f_\Theta$ is shared between Experience Replay Module and Feature Calibration Module to extract image representations, then an auxiliary classifier $f_\psi$  with a non-linear MLP and a classifier $f_\theta$  with a single linear layer are applied on top of the image representations to predict labels. A non-linear projector $f_\phi$ in Feature Calibration Module is adopted to translate the image representations for the supervised contrastive learning loss.}
	\label{fig:SC}
\end{figure*}

To tackle the above problems, we propose a Complementary Calibration (CoCa) framework to alleviate the relation and feature deviations by mining the complementary information of model's outputs and features. Particularly, we first utilize the collaborative distillation technique by ensembling dark knowledge to balance the relationship among classes while maintaining the performance of old tasks. Then, we employ collaborative self-supervision composed by pretext tasks and supervised contrastive learning, in which pretext tasks enable feature extractor to learn complete features, while supervised contrastive learning maintains the meaningful transformation of pretext tasks and learns discriminative features between the new and old classes. Pretext tasks and supervised contrastive learning complement each other, ensuring the feature representations to be complete and discriminative for all classes in GCL. The proposed framework is shown in Fig. \ref{fig:SC}. \par

The main highlights are summarized below:\par
(1) We reveal that relation and feature deviations are crucial problems for catastrophic forgetting in GCL, and  propose a novel Complementary Calibration (CoCa) framework for GCL to alleviate these two deviations by exploring the complementary information of model's outputs and  features.\par
(2) We ensemble dark knowledge to alleviate the relation deviation, keeping the performance of the old tasks and balancing the relationship of inter-class by  collaborative distillation.\par
(3) We leverage a collaborative self-supervised network by exploiting pretext tasks and supervised contrastive learning, which enables feature extractor to learn complete and discriminative features  for all classes in GCL.\par
(4) Extensive experiments on four popular datasets, namely sequential CIFAR-10, sequential CIFAR-100, sequential Tiny ImageNet and MNIST-360, show that our CoCa framework outperforms the state-of-the-art methods.

\section{Related work}
\subsection{Continual Learning}
Early studies focus on Task Incremental Learning (Task-IL) \cite{review_pami}, the simplest continual learning scenario, in which  the task boundaries are available during both training and testing stages. Such approaches can be roughly categorized into three types: rehearsal-based methods, regularization-based methods and structure-based methods. Rehearsal-based methods \cite{1995er},  \cite{GEM}, \cite{AGEM} replay reserved samples from old tasks while learning a new task to mitigate catastrophic forgetting. 
Regularization-based methods \cite{EWC}, \cite{SI}, \cite{oewc} constrain the parameters of each part of the model to protect the previously learned knowledge. Structure-based methods \cite{acl}, \cite{packnet}, \cite{hat} alleviate catastrophic forgetting by modifying the underlying architecture of the network. \par
Due to the rigid restriction of Task-IL, recent studies pay more attention on Class Incremental Learning (Class-IL) \cite{lucir}, \cite{icarl}, which prohibits access to the task boundaries during  testing stage. Different from Task-IL, Class-IL  needs to distinguish all seen classes when testing stage. Many studies \cite{lucir}, \cite{bic}, \cite{wa} have revealed that Class-IL models are easily biased  into new classes, thus existing efforts usually aim at alleviating this problem by  deviation amendment from different aspects, such as normalized features \cite{lucir}, class statistics \cite{IL2M} and weight aligning \cite{wa}. \par
Although great progress has been achieved, most Task-IL and Class-IL approaches depend on task boundaries in the training stage. Actually, task boundaries  are  blurry  in practical scenarios due to the fact that stream data usually have not clear task divisions. Thus, recent studies set out to explore General Continual Learning (GCL) \cite{review_pami}, \cite{NEURIPS2020_DER}, whose differences from the Task-IL and Class-IL settings are mainly in two aspects: (i) Task boundaries are not necessary during both training and testing stages; (ii) Memory size is limited even facing infinite stream data.  Therefore, it is a quite challenging setting. \par
Some efforts towards GCL are from the aspect of the sample strategy. Isele  \textit{et al.} \cite{ser} employed reservoir sample strategy  so that the probability of all samples can be selected is equal. Aljundi  \textit{et al.} \cite{gss} proposed a greedy selection method named Gradient based Sample Selection (GSS), which aims at improving the diversity of samples.  Afterwards, some methods concentrate on mixed methods. Rao \textit{et al.} \cite{curl} proposed a unsupervised continual learning approach called CURL with model expansion and generative replay to maintain the performance of old tasks. On the basis of CURL, Lee \textit{et al.} \cite{DPM} proposed Continual Neural Dirichlet Process Mixture (CN-DPM), which utilizes  the Bayesian nonparametric framework to enlarge the number of experts. Buzzega \textit{et al.} \cite{NEURIPS2020_DER} proposed Dark Experience Relay (DER) with the combination of regularization and rehearsal-based methods, which employs experience replay and knowledge distillation to promote the 
new model's outputs consistency with 
the original outputs. Our CoCa framewok also combines knowledge distillation with rehearsal, the key difference  is that we explore collaborative
distillation to balance the relationship among all classes by utilizing new model's outputs. \par
\subsection{Knowledge Distillation}
Knowledge distillation \cite{kd_review} refers to the approach that the training process of a student model is supported under the supervision of a teacher model with dark knowledge (soft targets). The dark knowledge contains the rich similarity relationship among all classes. In addition, the student model could distill knowledge by itself, which is called self-distillation \cite{ge2020bake}.   Knowledge distillation is widely applied in continual learning to address the catastrophic forgetting problem.  LWF \cite{LWF}  is the earliest work to explore it in continual learning, which aims at leveraging new samples of the old model's outputs to constrain  the new model's outputs. Afterwards, FDR \cite{FDR} stores samples as well as the dark knowledge of the old model, and constrains the $\ell_2$ norm of the difference between the new model's outputs and dark knowledge. Unlike the label distillation, LUCIR \cite{lucir}  directly limits the normalized features extracted by the new model as consistent as those by the old model, while PODNet \cite{podnet} constrains the evolution of each layer's output. Further, DDE \cite{dde} introduces causal inference to distill the casual effect between the old and new data. Different from them, our proposed collaborative distillation explores ensemble dark knowledge from old and new models, which contains more informative similarity relationships than that from a single model. \par
\subsection{Self-Supervised Learning}
Self-Supervised Learning (SSL) \cite{review_ssl} refers to learn representations with large amounts of data without manual labels, which explores input samples' inherent co-occurrence relationships as supervision. A typical type of SSL is pretext tasks, which  generally leverage the spatial structure or sequential relationships of  input images, such as pretext-invariant representations\cite{PIRL} and geometric transformation \cite{gidaris2018unsupervised}. Another type is unsupervised contrastive learning \cite{simclr}, \cite{understanding_contrastive}, which utilizes the contrastive loss to pull multiple views of an image closer and  pushes those from other samples apart in an embedding space.  SSL is widely applied in many fields, including few-shot learning \cite{EQINV}, imbalance learning \cite{cpsc} and continual learning \cite{pass}. Zhu  \textit{et al.} \cite{pass}  utilized  SSL to extract discriminative feature representations and memorized class-representative prototype  to maintain the class boundaries.  Khosla \textit{et al.} \cite{NEURIPS2020_scl} extended unsupervised contrastive learning to supervised setting by employing images from the same class as positive samples. Positive samples are pulled closer and the other samples are pushed away in an embedding space. Mai \textit{et al.} \cite{scr} proposed Supervised Contrastive Replay (SCR),  leveraging supervised contrastive learning and nearest-class-mean  classifier to mitigate catastrophic forgetting.\par
The closest studies to our work are  PASS \cite{pass} and SCR \cite{scr}. Particularly, PASS  memorizes each class-representative prototype that depends on task boundaries.  SCR \cite{scr}  adopts supervised contrastive learning to obtain well-separated representations, in which the trainings of feature extractor and classifier are separated. Moreover, the samples of old classes are apt to be lost, especially when the memory is very limited. Thus, SCR is hard to be applied in GCL. Different from them, we introduce pretext tasks and supervised contrastive learning into collaborative self-supervision in our GCL framework, which complement each other in our Feature Calibration Module. In this way, complete and discriminative features can be obtained. \par
\section{Methodology}
In this work, we focus on GCL in image classification. Formally, given  stream data  characterized by tasks $\mathcal{D}_1,\mathcal{D}_2,\dots ,\mathcal{D}_T$, each task $\mathcal{D}_t=\{\left( x_{t,i},y_{t,i} \right) _{i=1}^{N_t}\}$  consists of $N_t$ images $x\in \mathcal{X}$  and labels $y\in \mathcal{Y} $  corresponding to $x$. The task boundaries are unavailable during both training and testing stages. To alleviate catastrophic forgetting, similar to ER \cite{ser}, we employ constant memory $\mathcal{M}=\{\left( x_i,\hat{o}_i,y_i \right) _{i=1}^{\mathcal{B}}\}$ to store limited amount of previous training samples $(x,y)$ and corresponding model's outputs $\hat{o}$, where $\mathcal{B}$ represents buffer size. \par
As shown in Fig. \ref{fig:SC}, our proposed CoCa framework consists of three parts: an Experience Replay Module, a Relation
Calibration Module and a Feature Calibration Module. In what follows, we first introduce Experience Replay Module and analyze  the relation deviation among classes in existing knowledge distillation methods. Then in Relation Calibration Module,  we elaborate on how collaborative knowledge distillation mitigates the relation deviation.  Finally, the Feature Calibration Module leverages collaborative self-supervision to alleviate the feature  deviation, which is composed of pretext tasks based on geometric transformations and supervised contrastive learning to learn complete and discriminative features respectively. Interestingly, the alleviation process for relation and feature deviations are mutually complementary.  \par
\subsection{Experience Replay Module}
Our feature extractor $f_\Theta$ for GCL is a neural network, parameterized with parameters $\Theta$. The role of $f_\Theta$ is to extract feature representations of all tasks. Meanwhile, a classifier $f_\theta$ needs to be trained to project the learned feature representations into the label space. The ideal goal is to minimize the following formula:
\begin{equation}
\underset{\Theta, \theta}{\arg \min } \sum_{t=1}^{T} \mathbb{E}_{(x, y) \sim \mathcal{D}_{t}} \mathcal{L}_{C E}\left(\sigma\left(f_{\Theta, \theta}(x)\right), y\right),
\label{eq1}
\end{equation}
where $\mathcal{L}_{CE}$ is the cross-entropy loss and $\sigma ( \cdot ) $ is the softmax function. Since it is unavailable to get all samples from old tasks, Experience Reply Module stores a subset of previous training set and employs them to jointly optimize models. This is equivalent  to correctly classifying new tasks given limited memory $\mathcal{M}$ from old tasks. Therefore, Eq. \ref{eq1} is substituted by the following loss terms:
\begin{equation}
\begin{aligned}
\mathcal{L}_{\text {base }} &=\mathbb{E}_{(x, y) \sim \mathcal{D}_{t}} \mathcal{L}_{C E}\left(\sigma\left(f_{\Theta, \theta}(x)\right), y\right) \\
&+\mathbb{E}_{(x, y) \sim \mathcal{M}} \mathcal{L}_{C E}\left(\sigma\left(f_{\Theta, \theta}(x)\right), y\right).
\end{aligned}
\label{eq2}
\end{equation}
 \par
Experience Replay Module is widely applied in image classification. For example, Buzzega \textit{et al.} \cite{NEURIPS2020_DER} proposed a strong baseline dubbed DER that employs experience replay and knowledge distillation to maintain the performance of old tasks on GCL setting. In particular, it constrains the new model's outputs to be as consistent as possible with the old model. Formally, the loss is written as:
\begin{equation}
\mathcal{L}_{K D}=\mathbb{E}_{(x, \hat{o}) \sim \mathcal{M}}\left\|f_{\Theta, \theta}(x)-\hat{o}\right\|_{2}^{2},
\label{eq3}
\end{equation}
where $\hat{o}$ are the model's outputs sampled from old models and the $\lVert \cdot \rVert _2$ operator refers to the $\ell_2$ norm.\par
However, when training the new model in GCL, $\hat{o}$ is 
deficient to express the relationship among all classes. We refer to this deficiency as relation deviation. 
\subsection{Relation Calibration Module}
As mentioned above, the relation deviation in GCL is caused by forcing the outputs of new model to keep consistent with the old model's outputs, which lack whole inter-class relationships compared with the outputs of new model. Meanwhile, when the model is trained on a new task, its performance on old tasks usually drops significantly, indicating inaccurate outputs of the new model on old tasks. Fortunately, this deficient inter-class relationships exist in the new model's outputs and this inaccuracy can be compensated by the old model's outputs. Therefore,  we naturally aim to explore complementary properties of the two outputs by forcing the new model's outputs to be consistent with the  ensemble outputs of old and new  models. \par 
Specifically, we propose a collaborative distillation loss $\mathcal{L}_{CKD}$ to keep the new model’s outputs $o$ consistent with the ensemble outputs $o^*(o, \hat{o})$ obtained by Relation Calibration Module. 
It utilizes the features similarity matrix, the outputs $o$ of the new model and the reserved outputs  $\hat{o}$  of the old model to obtain ensemble outputs $o^*(o, \hat{o})$. Hence, the learning objective for collaborative knowledge distillation is:
\begin{equation}
	\mathcal{L}_{CKD}= \mathbb{E}_{\left( x,\hat{o} \right) \sim\mathcal{M}}\lVert o -o^*(o, \hat{o}) \rVert _{2}^{2}.
	 \label{eq4}
\end{equation} \par
Noticeably, we cannot directly combine $o$ and $\hat{o}$ linearly. Otherwise, if the ensemble outputs $o^*$ are composed by $o^*(o, \hat{o})=\gamma o+\left( 1-\gamma \right) \hat{o}$, where $\gamma$ is a positive trade-off hyper-parameter to
balance $\hat{o}$ and $o$, $\mathcal{L}_{C K D}$ and $\mathcal{L}_{K D}$ will be linearly correlated: 
\begin{equation} \label{eq5}
\begin{aligned}
\mathcal{L}_{CKD} &=\mathbb{E}_{( x,\hat{o} ) \sim \mathcal{M}}\lVert o-( \gamma o+ ( 1-\gamma ) \hat{o} ) \rVert _{2}^{2} \\
&= ( 1-\gamma ) ^2\mathcal{L}_{KD}.
\end{aligned}
\end{equation} \par
Meanwhile, the new model's outputs may inaccurate. Therefore, we adopt the features similarity in the same batch to propagate the new model’s outputs $o$ and fuse the old model’s outputs $\hat{o}$.  In this way, the ensemble outputs $o^*(o, \hat{o})$ are carried out to obtain complete inter-class relationships, which are calculated by the following three steps.
Firstly, we calculate the normalized samples’ similarity $\hat{\mathcal{S}}(i, j)$ in the same batch. For each pair of sample $\left( x_i,x_j \right)$, the normalized  feature embeddings obtained by the feature extractor $f_{\Theta}$  is $\left( \hat{\bold{z}}_i,\hat{\bold{z}}_j \right)$, the normalized samples’  similarity matrix $\hat{\mathcal{S}}\in {R}^{N\times N}$ is calculated by:
\begin{equation}
\hat{\mathcal{S}}(i, j)=\frac{\exp (\mathcal{S}(i, j))}{\Sigma_{k \neq i} \exp (\mathcal{S}(i, k))},
\label{eq6}
\end{equation}
where the similarity function $\mathcal{S}(i,j)$  is defined as:
\begin{equation}
\mathcal{S}(i, j)=\left\{\begin{array}{l}
\hat{\bold{z}}_{i}^{T} \hat{\bold{z}}_{j}(i \neq j) \\
0(i=j)
\end{array}\right..
\label{eq7}
\end{equation}\par
Secondly, we conduct label propagation by $o$ and normalized similarity matrix $\hat{\mathcal{S}}$ described in \cite{NIPS2003_87682805} as:
\begin{equation}
	Q_t=\omega \hat{\mathcal{S}}Q_{t-1}+\left( 1-\omega \right) o\left( t\ge 0 \right),
	\label{eq8}
\end{equation}
where $\omega$ is a weighting factor in range [0, 1) and $Q_0 = o$. The label propagation is conducted many times for obtaining more accurate outputs:
\begin{equation}
	\begin{split}\label{eq9}
	Q_{\infty}=\lim_{t\rightarrow \infty}\left[ \left( \omega \hat{\mathcal{S}} \right) ^t o+\left( 1-\omega \right) \sum_{i=1}^{t-1}{\left( \omega \hat{\mathcal{S}} \right)}^i o \right] .	
	\end{split}
\end{equation}\par
Since $\omega$ and the eigenvalues of $\hat{\mathcal{S}}$ are both in range [0, 1), we obtain an approximate formulation for $Q_\infty$ as:
\begin{equation}
	\begin{split}\label{eq10}
	Q_{\infty}=\left( 1-\omega \right) \left( I-\omega \hat{\mathcal{S}} \right) ^{-1}o,	
	\end{split}
\end{equation}
where $I$ is an identity matrix.\par
Finally, the ensemble outputs $o^*$ consist of the old model’s outputs $\hat{o}$ and the modified outputs $Q_\infty$, which is written as:
\begin{equation}
	o^*(o, \hat{o})=\gamma \left( 1-\omega \right) \left( I-\omega \hat{\mathcal{S}} \right) ^{-1}o+\left( 1-\gamma \right) \hat{o}.
	\label{eq11}
\end{equation} \par
In this way, we alleviate the relation deviation among all classes of the old model's outputs. Even if the matrix inversion operation takes $O(n^3)$ time complexity,  the computational complexity is trivial when the batch size $n$ is limited. \par
\subsection{Feature Calibration Module}
Besides relation deviation, feature deviation is another key challenge, which is caused by the indiscriminative feature representations. To address this challenge, we develop a Feature Calibration Module, which consists of  pretext tasks and supervised contrastive learning. In detail, we first design self-supervised pretext tasks as auxiliary supervision, enabling feature extractor to learn complete features. Then, we utilize the supervised contrastive learning to learn discriminative features between the new and old classes. \par
As the samples of the old tasks are unavailable in GCL, the  feature extractor tends to  extract the discriminative features for the new incoming task. This tendency results in incomplete feature representations, which generally cannot well distinguish the old tasks from the new ones. To calibrate incomplete feature representations, we exploit self-supervised learning based on  geometric transformations pretext tasks to enable the feature extractor to exact complete features, which represents the rich spatial or sequential relationships of the samples. In particalr, we apply pretext tasks loss $\mathcal{L}_{PT}$ to distinguish what geometric transformation has been made to the original images. Among them, a muti-layers perception $f_\psi$ is applied as the  auxiliary  classifier to project the feature exacted by $f_\Theta$ into the label space.  Accordingly, the pretext tasks loss $\mathcal{L}_{PT}$  of geometric transformation tasks are designed as:
\begin{equation}\label{eq12}
	\mathcal{L}_{PT} =\mathcal{L}_{CE}( \sigma ( f_{\Theta ,\psi}\left( x^p ,y^p \right) )) ,
\end{equation}
where the proxy label $y^p$ consists of a series of geometric transformations, such as rotation and scaling, image $x^p$ is produced by applying geometric transformations proxy label $y^p$ to the original image $x$.\par
To better learn discriminative features for all tasks, we further leverage supervised contrastive learning in CoCa framework. In detail, we introduce another MLP $f_\phi$ with parameters $\phi$, whose purpose is to map the feature to an embedding space where the supervised contrastive learning loss is applied. In the feature space, the distances from the same class are shorten and those of different classes are enlarged. Assuming the embedding $\bold{z}=f_{\Theta ,\phi}( x ) $, $\{\bold{z}_{i}^{+}\} $ and $\{\bold{z}_{i}^{-}\}$ represent the set of all positive and negative samples distinct from $\bold{z}_i$ in the multi-viewed batch respectively, the supervised contrastive learning loss function is written as:
\begin{equation}
\begin{array}{l}
\mathcal{L}_{S C L}=\mathbb{E}_{\bold{z}_{i} \sim \bold{z}}[-\underset{\bold{z}_{j} \sim \{\bold{z}_{i}^{+}\}}{\Sigma} \frac{\mathcal{S}(\bold{z}_{i}, \bold{z}_{j})}{\tau}+ \\
\log (\underset{\bold{z}_{j} \sim \{\bold{z}_{i}^{+}\}}{\Sigma} exp({ \frac{\mathcal{S}(\bold{z}_{i}, \bold{z}_{j})}{\tau} })+ \underset{\bold{z}_{k} \sim \{\bold{z}_{i}^{-}\}}{\Sigma} exp(\frac{\mathcal{S}(\bold{z}_{i}, \bold{z}_{k})}{\tau}))], 
\end{array}
\label{eq13}
\end{equation}
where $(i,j,k)$ denotes the index, $\tau$ is a scalar temperature parameter, and $\underset{\bold{z}_{k} \sim \{\bold{z}_{i}^{-}\}}{\Sigma} exp(\frac{\mathcal{S}(\bold{z}_{i}, \bold{z}_{k})}{\tau})$ encourages samples from different classes to be as far away from the unit hypersphere as possible in Eq. \ref{eq13}. In this way, the features are evenly distributed on the unit hypersphere as far as possible.  Supervised contrastive learning could well distinguish the old and new classes as it utilizes the label information.\par
Noticeably, pretext tasks and supervised contrastive learning are  cooperative and complementary. Here, they not only cooperate together to obtain better feature representations to alleviate the feature deviation problem, but make up for their respective defects mutually. Concretely, on the one hand, the pretext tasks may be redundant, and even may produce interference. For example, it is difficult to distinguish number 6 from the number 9 after $180^{\circ}$ rotation. Thankfully, the supervised contrastive learning suppresses the redundancy of the pretext tasks by supervised information; on the other hand,  when  the samples of the old classes are missing due to the limited memory, the complete features learned by the pretext tasks assist the supervised contrastive learning to obtain discriminative features for all tasks. Therefore, the collaborative self-supervision loss $\mathcal{L}_{CSS}$ of Feature Calibration Module becomes:
\begin{equation}
\mathcal{L}_{CSS}= \mathcal{L}_{PT}+ \mathcal{L}_{S C L}.
\label{eq14}
\end{equation} \par 
Feature Calibration Module explores existing samples and features to learn complete and discriminative features for both old and new tasks, these complementary sets of features alleviate catastrophic forgetting effectively.
\subsection{The Overall Objective }
Since our proposed CoCa framework consists of Experience Replay Module, Feature Calibration Module and Relation Calibration Module,  the objective function of the whole training stage is as follows:
\begin{equation}
	\mathcal{L}_{CoCa}=\mathcal{L}_{base}+\lambda_1 \mathcal{L}_{CKD}+\lambda_2 \mathcal{L}_{CSS}, \label{eq15}
\end{equation}
where $\lambda_1$ and $\lambda_2 $ are hyperparameters. At the test stage, the Feature Calibration Module and Relation Calibration Module are removed.
\section{Experiments}
\begin{table*}[]
\caption{Average accuracy (\%) on GCL setting. “*” indicates that task boundaries are provided at the training stage} \label{class}
\begin{center}
\begin{tabular}{ccccccccccccc}
\toprule [1 pt]
Method & \multicolumn{3}{c}{sequential CIFAR-10}     &\multicolumn{3}{c}{sequential CIFAR-100} &\multicolumn{3}{c}{sequential Tiny ImageNet} &\multicolumn{3}{c}{MNIST-360}\\ \hline
JOINT  & \multicolumn{3}{c}{92.20}  & \multicolumn{3}{c}{69.55} & \multicolumn{3}{c}{59.99}  & \multicolumn{3}{c}{82.98}\\
SGD & \multicolumn{3}{c}{19.62} & \multicolumn{3}{c}{4.33}  & \multicolumn{3}{c}{7.92} & \multicolumn{3}{c}{19.02}\\ \hline
LWF*\cite{LWF}   & \multicolumn{3}{c}{19.61}  & \multicolumn{3}{c}{4.26} & \multicolumn{3}{c}{8.46} & \multicolumn{3}{c}{-}\\
oEWC*\cite{oewc}   & \multicolumn{3}{c}{19.49}  & \multicolumn{3}{c}{3.49} & \multicolumn{3}{c}{7.58} & \multicolumn{3}{c}{-} \\
SI*\cite{SI}     & \multicolumn{3}{c}{19.48}  & \multicolumn{3}{c}{4.60} & \multicolumn{3}{c}{6.58} & \multicolumn{3}{c}{-}\\
CN-DPM\cite{DPM} & \multicolumn{3}{c}{45.21}  & \multicolumn{3}{c}{20.10} & \multicolumn{3}{c}{-} & \multicolumn{3}{c}{-} \\ \hline
$\mathcal{B}$ & 200 & 500 & 5120 & 200 & 500 & 5120  & 200 & 500 & 5120 & 200 & 500 & 1000 \\ \hline
GEM*\cite{GEM} & 25.54  & 26.20 & 25.26 & 8.17 & 12.45  & 4.55  &{-}  &{-}  &{-} &{-}  &{-}  &{-} \\
iCaRL*\cite{icarl} & 49.02 & 47.55  & 55.07 & 19.26  & 24.71 & 29.78 & 7.53  & 9.38  & 14.08 &{-}  &{-}  &{-}\\
FDR*\cite{FDR} & 30.91 & 28.71 & 19.70 & 11.67 & 18.00 & 32.35 & 8.70 & 10.54 & 28.97  &{-}  &{-}  &{-}\\
HAL*\cite{HAL}  & 32.36  & 41.79 & 59.12  & 7.60  & 9.55 & 22.11 &{-}  &{-}  &{-} &{-}  &{-}  &{-} \\
ER\cite{ser}  & 44.79 & 57.74 & 82.47 & 9.84 & 14.64 & 44.79 & 8.49 & 9.99 & 27.40 & 49.27 & 65.04 & 75.18  \\
A-GEM\cite{AGEM} & 20.04 & 22.67  & 21.99 & 4.73   & 4.74   & 4.87 & 8.07  & 8.06   & 7.96 & 28.34& 28.13 & 29.21 \\
GSS \cite{gss} & 39.07  & 49.73  & 67.27 & 6.35 & 7.44 & 9.71   &{-}  &{-}  &{-}  & 43.92 & 54.45 & 63.84\\
DER\cite{NEURIPS2020_DER} & 64.88  & 72.70  & 85.24 & 18.66  & 28.70  & 51.20 & 10.96  & 19.38  & 39.02 & 54.16 & 69.62 & 76.03\\
\textbf{CoCa (ours)}  & \textbf{66.25} & \textbf{76.27} & \textbf{89.52} & \textbf{21.20} & \textbf{32.88} & \textbf{58.38}  & \textbf{12.78} & \textbf{20.33} & \textbf{39.79} & \textbf{67.02} & \textbf{76.49}  & \textbf{81.82} \\ \toprule [1 pt]
\end{tabular}
\end{center}
\end{table*}
\begin{table*}[]
\caption{Average accuracy (\%) on Task-IL setting.  “*” indicates that task boundaries are provided at the training stage} \label{task}
\begin{center}
\begin{tabular}{cccccccccc}
\toprule [1 pt]
Method & \multicolumn{3}{c}{sequential CIFAR-10}     &\multicolumn{3}{c}{sequential CIFAR-100} &\multicolumn{3}{c}{sequential Tiny ImageNet} \\ \hline
JOINT    & \multicolumn{3}{c}{98.31}   & \multicolumn{3}{c}{95.22}  & \multicolumn{3}{c}{82.04}  \\
SGD      & \multicolumn{3}{c}{61.02}  & \multicolumn{3}{c}{39.91}& \multicolumn{3}{c}{18.31}\\ \hline
LWF*\cite{LWF}   & \multicolumn{3}{c}{63.29}     & \multicolumn{3}{c}{25.02}& \multicolumn{3}{c}{15.85}\\
oEWC*\cite{oewc}   & \multicolumn{3}{c}{68.29}    & \multicolumn{3}{c}{23.13} & \multicolumn{3}{c}{19.20}\\
SI*\cite{SI}    & \multicolumn{3}{c}{68.05}     & \multicolumn{3}{c}{38.37} & \multicolumn{3}{c}{36.32}\\
\hline
$\mathcal{B}$ & 200 & 500 & 5120 & 200 & 500 & 5120  & 200 & 500 & 5120 \\ \hline
GEM*\cite{GEM} & 90.44  & 92.16   & 95.55     & 68.56     & 71.69          & 78.21  &{-}  &{-}  &{-} \\
iCaRL*\cite{icarl}   & 88.99   & 88.22   & 92.23  & 71.56  & 77.99  & 81.59  & 28.19    & 31.55   & 40.83   \\
FDR*\cite{FDR}   & 91.01 & 93.29   & 94.32       & 68.78    & 75.20    & 82.33   & 40.36    & 49.88     & 68.01 \\
HAL*\cite{HAL}   & 82.51  & 84.54  & 88.51            & 51.49          & 57.03          & 67.42 &{-}  &{-}  &{-} \\
ER\cite{ser}    & 91.19   & 93.61   & 96.98    & 69.40    & 74.45    & 88.83     & 38.17   & 48.64    & 67.29  \\
A-GEM\cite{AGEM}  & 83.88    & 89.48   & 90.10 & 58.88   & 59.58    & 64.68 & 22.77    & 25.33    & 26.22   \\
GSS \cite{gss}   & 88.80   & 91.02   & 94.19          & 41.94          & 56.18          & 67.65    &{-}  &{-}  &{-} \\
DER\cite{NEURIPS2020_DER}  & 91.92   & 93.88  & 96.12  & 72.21   & 77.30   & 87.66& 40.87          & 51.91          & 69.84\\
\textbf{CoCa (ours)}  & \textbf{92.95} & \textbf{94.54} & \textbf{97.17} & \textbf{75.23} & \textbf{82.28} & \textbf{91.36}& \textbf{46.13} & \textbf{55.03} & \textbf{71.19}  \\ \toprule [1 pt]
\end{tabular}
\end{center}
\end{table*}
\subsection{Datasets} 
Four popular datasets are selected  in our image classification experiments: sequential CIFAR-10, sequential CIFAR-100, sequential Tiny ImageNet and MNIST-360. \par
CIFAR-10 \cite{2009CIFAR} consists of 10 classes, each class has 6000 samples of 32$\times$32 color images, including 5000 training samples and 1000 test samples. CIFAR-100 \cite{2009CIFAR} is similar to CIFAR-10 except that the number of classes is 100. Each class has 600 images, which are divided into 500 for training and 100 for testing. Tiny ImageNet \cite{0Tiny} is a subset of ImageNet \cite{imagenet}, which contains 200 classes, and each class has 500 samples with 64 $\times$ 64 color images for training. We split the CIFAR-10, CIFAR-100 and Tiny ImageNet evenly into 5, 20 and 10 sequential tasks respectively, each of which includes 2, 5 and 20 classes, 
i.e., sequential CIFAR-10, sequential CIFAR-100 and sequential Tiny ImageNet.  MNIST-360 \cite{NEURIPS2020_DER} is specially designed for GCL setting, which offers a sequence of MNIST numbers from 0 to 8 at increasing angles. It builds the batch by using samples belong to two continuous subsequent classes at a time, such as, (0, 1), (1, 2), ... , (8, 0). Different from other three datasets, the task boundaries are blurry and the same task appears repeatedly. It is more challenging and practical for continual learning. \par
\subsection{Implementation details}  
We employ a fully connected network with two hidden layers as backbone for MNIST-360 dataset. As for the other datasets, we employ ResNet-18 \cite{Resnet}  as backbone. And two fully connected networks with  three  hidden layers are employed as the  projector and auxiliary classifier, respectively. The hyperparameters are selected via grid search by employing reserved samples of validation set from all task's training sets. The number of epochs for datasets MNIST-360, sequential CIFAR-10, sequential CIFAR-100 and sequential Tiny ImageNet are 1, 50, 50 and 100, respectively.  Following \cite{NEURIPS2020_DER}, we utilize
Stochastic Gradient Descent (SGD) as optimizer and  fix the batch size at 64 to ensure that the amount of updates for all methods are the same. Concretely, each batch consists of 32 new samples and 32 reserved samples, and the later are updated with reservoir sample strategy \cite{reservoir1985}  at the end of each batch.\par
In Relation Calibration Module, we set both the weighting factor $\omega$ and the trade-off hyper-parameter $\gamma$ as 0.1. In Feature Calibration Module, the  pretext tasks are composed of three types of geometric transformations, including rotation $ \{0^{\circ}
, 90^{\circ}, 180^{\circ}, 270^{\circ}\}$, scaling $\{ 0.67, 1.0 \}$
 and aspect ratio $ \{ 0.67, 1.33 \}$. Additionally, we only perform a random transformation for a sample to save the computation cost. Our approach is implemented with pytorch framework and trained on one NVIDIA GeForce RTX 3060 GPU.\par
\subsection{Comparison with State-of-the-Art Methods}
\textit{1) Competitors:}  Two group of competitors are selected.  The first group is the GCL approaches, including (1) CN-DPM (continual neural dirichlet process mixture) \cite{DPM}; (2) ER (experience replay with reservoir) \cite{ser}; (3) A-GEM
(average gradient episode memory) \cite{AGEM}; (4) GSS (gradient sample selection) \cite{gss}; and (5) DER (dark experience replay) \cite{NEURIPS2020_DER}. The second group is the Class-IL and Task-IL approaches, in which task boundaries should be provided during training stage, including: (1) LWF (learning without forgetting) \cite{LWF}; (2) oEWC (online elastic weight consolidation) \cite{oewc}; (3) SI (synaptic intelligence) \cite{SI}; (4) GEM (gradient episode memory) \cite{GEM};  (5) iCaRL (incremental classifier and representation learning) \cite{icarl}; (6) FDR (function distance regularization) \cite{FDR}; and (7) HAL (hindsight anchor learning) \cite{HAL}. In addition, we also report the performance bound, including: (1) JOINT represents that all data are available at any time, which is an upper bound; and (2) SGD means that no strategy is adopted to alleviate forgetting at the training time, which is the lower bound. \par
Following \cite{review_pami} and \cite{NEURIPS2020_DER}, we employ the average accuracy on all tasks as the evaluation
criterion. To make a fair comparison, we apply the single-head setting during training stage and the pretrain model is unavailable in all methods. It should be emphasized that no task boundaries are provided for our proposed CoCa approach, even in comparison with those Class-IL and Task-IL approaches.\par
\textit{2) Results: } Table \ref{class} reports the results of the comparison on GCL setting. It could be observed that our approach outperforms the competitors in all cases. Specifically, for sequential CIFAR-10 dataset, the proposed CoCa framework outperforms the state-of-the-art competitors at least in 1.3\%. Furthermore, CoCa  beats  all methods on sequential CIFAR-100 dataset. For example, it surpasses the second-best method DER in around 7\% in buffer size sets of 500. As for the sequential Tiny ImageNet dataset,  CoCa has a performance gain against the best competitor by 1.82\%, 0.95\% and 0.77\% under 200, 500 and 5120 buffers, respectively. \par
Since the MNIST-360 dataset has no task boundaries, many methods that rely on task boundaries are inadequate for it, such as LWF \cite{LWF}, oEWC \cite{oewc}, SI \cite{SI}, GEM \cite{GEM}, iCaRL \cite{icarl}, FDR \cite{FDR}, and HAL \cite{HAL}. Thus, their methods are unavailable on MNIST-360 dataset. We could observe that our CoCa framework achieves top performance across all the different buffer sizes, which surpasses the suboptimal method at least in 5\%. It is quite impressive when the buffer size is limited, such as 200, with at least 12\% gains against the other competitors. These results indicate that the collaborative distillation and self-supervision  greatly alleviate catastrophic forgetting by mitigating deviation in the absence of task boundaries. \par
In addition, we also observe significant performance differences of  CoCa on different datasets and different buffer sizes. Specifically, it can be observed that the performance on sequential CIFAR-10 dataset is higher than sequential Tiny ImageNet and sequential CIFAR-100 datasets. The reason lies in that it has quite fewer classes than the other datasets.
For the  performance  differences among different buffer sizes for the same dataset, it is due to that the more replay samples are retained, the easier it is to alleviate the deviation. When the buffer size is large enough, it approximates the setting of joint training, i.e., the performance upper bound. As we can see, as the buffer size  increases from 200 to 5120, the average accuracy improves by at least 23\% on these datasets. \par
Moreover, since all methods are applicable to with Task-IL setting, following \cite{NEURIPS2020_DER}, we also make a comparison on Task-IL setting, as shown in Table \ref{task}. It should be noted that Task-IL experiments cannot be conducted on MNIST-360 dataset since there is no task boundaries during testing stage. We could observe that our proposed approach also outperforms all the competitors on the three datasets, even additional boundaries are provided for those Class-IL and Task-IL approaches. Take the sequential CIFAR-100 dataset  for example, the accuracies of our CoCa are 75.23\%, 82.28\% and 91.36\%  under 200, 500 and 5120 buffers, respectively.\par
\begin{figure*}[htbp]
	\centering
	\begin{center}
		\subfigure[$\omega$]{\label{fig:omega}
			\includegraphics[scale=0.51,trim=0 0 0 0,clip]{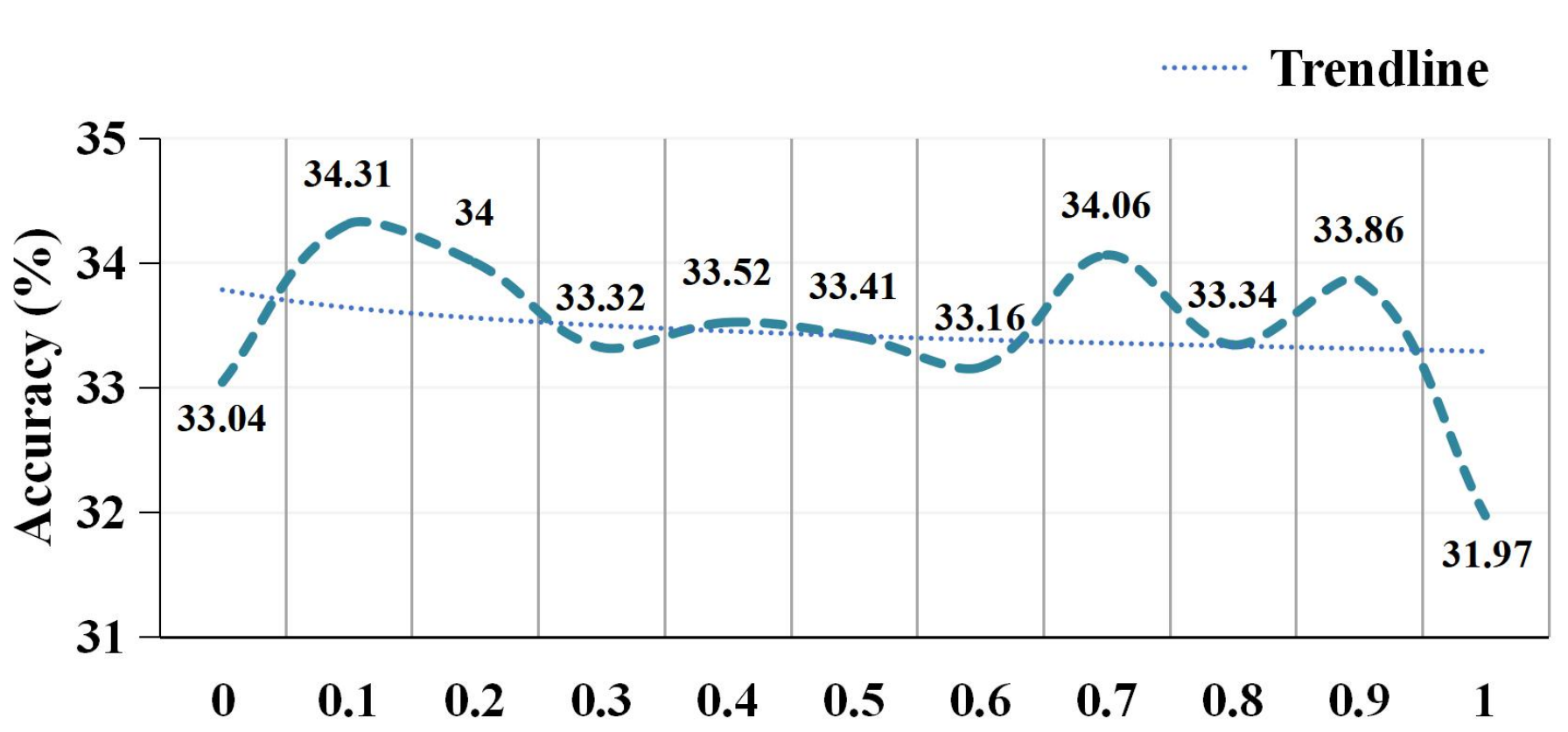}}
		\subfigure[$\gamma$]{\label{fig:gamma}
			\includegraphics[scale=0.45,trim=0 0 0 0,clip]{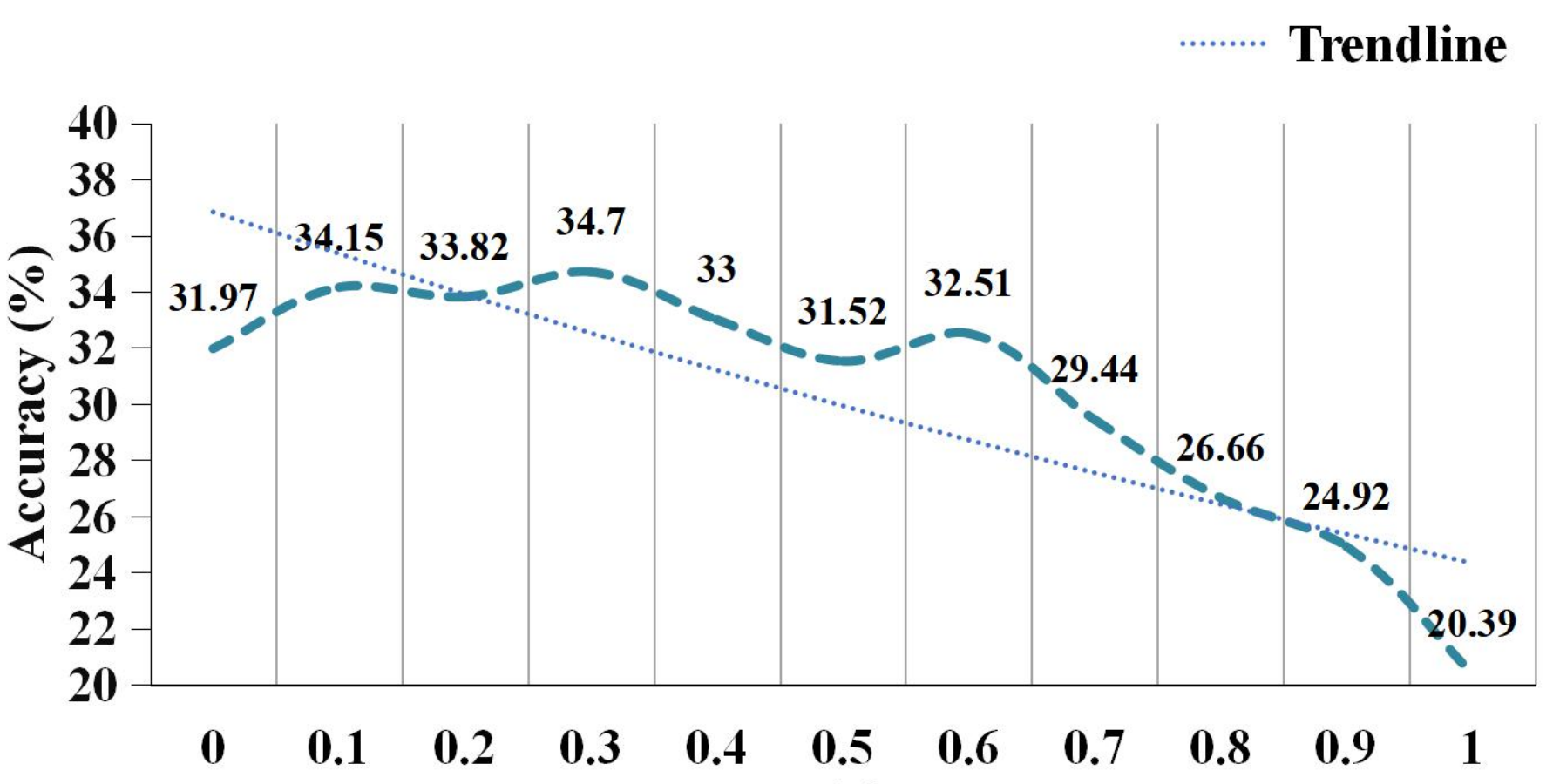}}
	\end{center}
	\caption{The impacts of relation calibration parameters $\omega$ and $\gamma$ on sequential CIFAR-100 dataset with buffer size of 500.}
	\label{fig:sensitivity}
\end{figure*}
\begin{figure*}[htbp]
	\centering
	\begin{center}
		\subfigure[SGD]{\label{fig:sgdt}
			\includegraphics[scale=0.4,trim=10 20 110 20,clip]{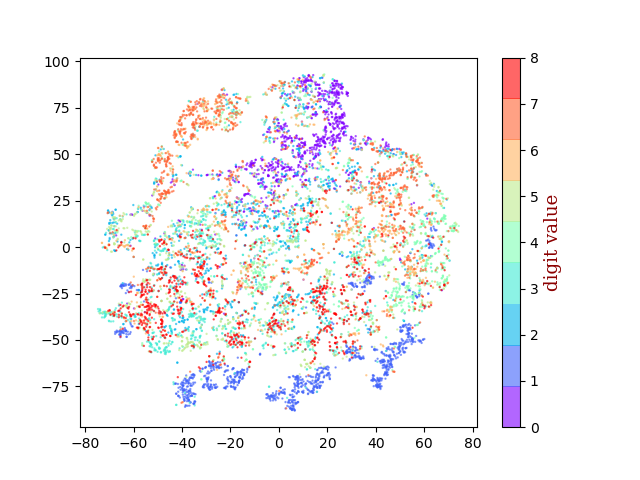}}
		\subfigure[ER]{\label{fig:ert}
			\includegraphics[scale=0.4,trim=10 20 110 20,clip]{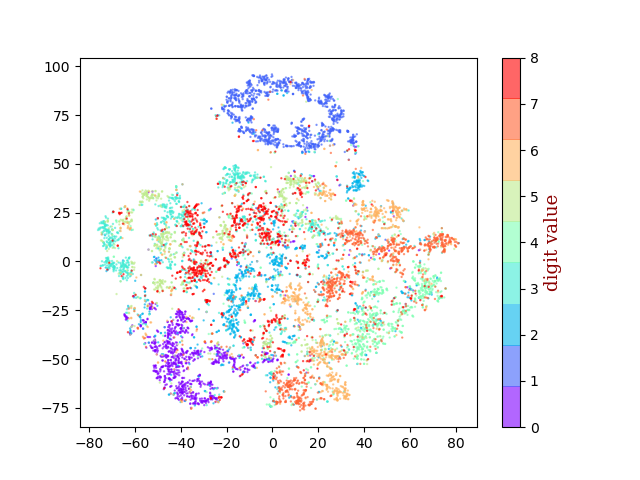}}
		\subfigure[Relation Calibration Module]{\label{fig:rct}
			\includegraphics[scale=0.4,trim=10 20 40 20,clip]{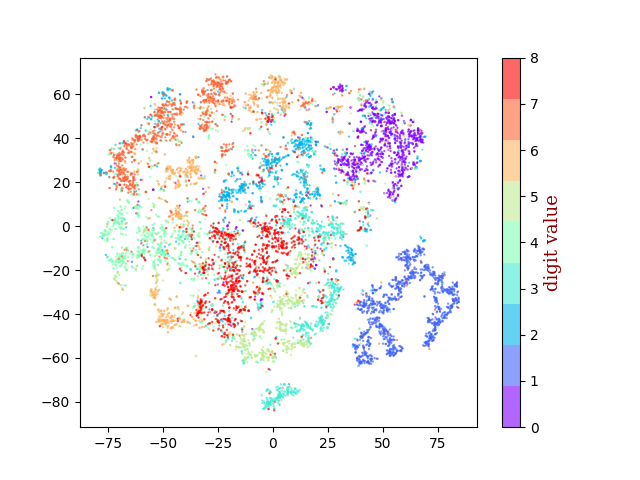}}
		\subfigure[JOINT]{\label{fig:jointt}
			\includegraphics[scale=0.4,trim=10 20 110 20,clip]{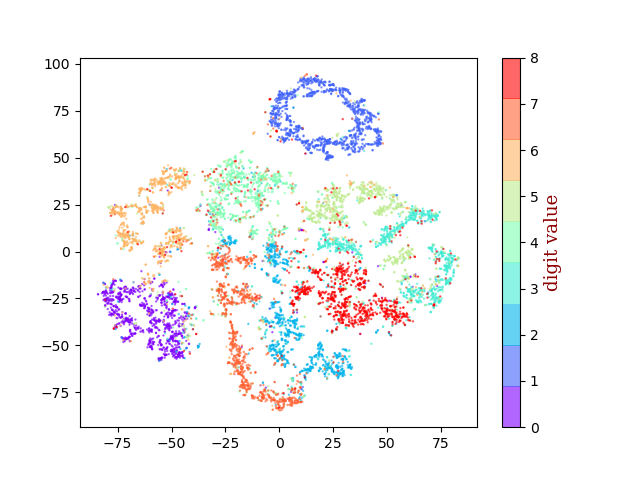}}
		\subfigure[Feature Calibration Module]{\label{fig:fct}
			\includegraphics[scale=0.4,trim=10 20 110 20,clip]{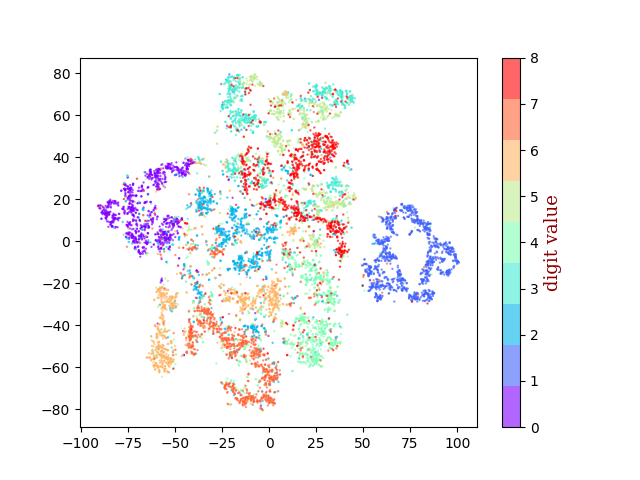}}
		\subfigure[\textbf{CoCa (ours)}]{\label{fig:sct}
			\includegraphics[scale=0.4,trim=10 20 40 20,clip]{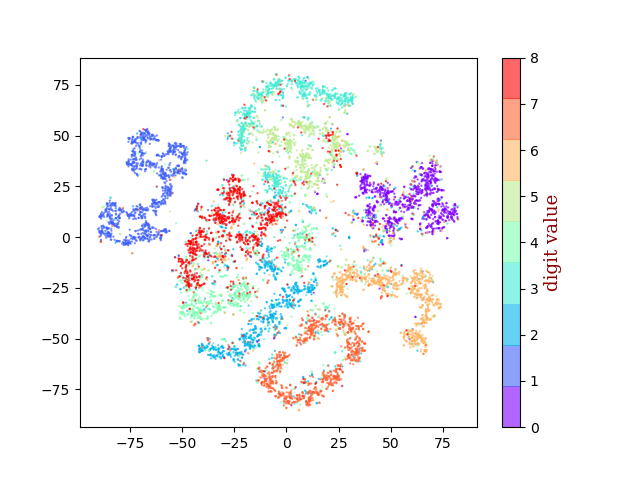}}
	\end{center}
	\caption{t-SNE visualization of features for test set of MNIST-360 dataset. (a) t-SNE visualization results of SGD. (b) t-SNE visualization results of ER. (c) t-SNE visualization results of Relation Calibration Module. (d) t-SNE visualization results of JOINT. (e) t-SNE visualization results of Feature Calibration Module. (f) t-SNE visualization results of CoCa.}
	\label{fig:tsne}
\end{figure*}
\begin{figure*}[htbp]
	\centering
	\begin{center}
		\subfigure[JOINT]{\label{fig:jointcf}
			\includegraphics[scale=0.33,trim=10 0 110 20,clip]{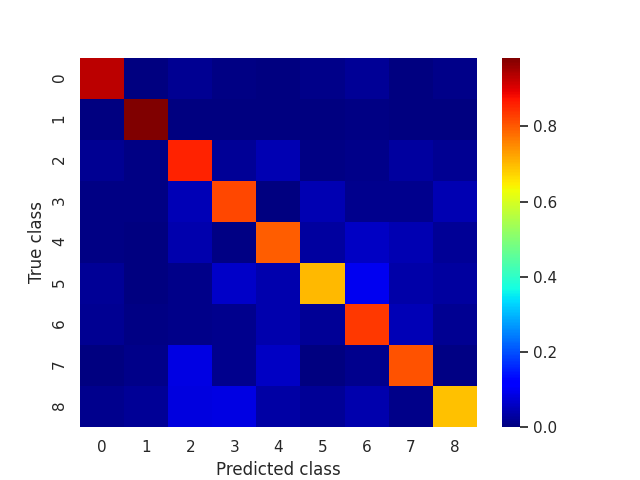}}
		\subfigure[SGD]{\label{fig:sgdcf}
			\includegraphics[scale=0.33,trim=10 0 110 20,clip]{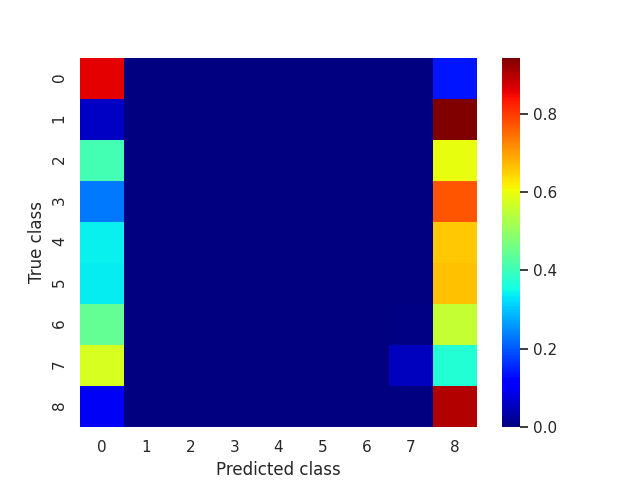}}
		\subfigure[ER]{\label{fig:ercf}
			\includegraphics[scale=0.33,trim=10 0 110 20,clip]{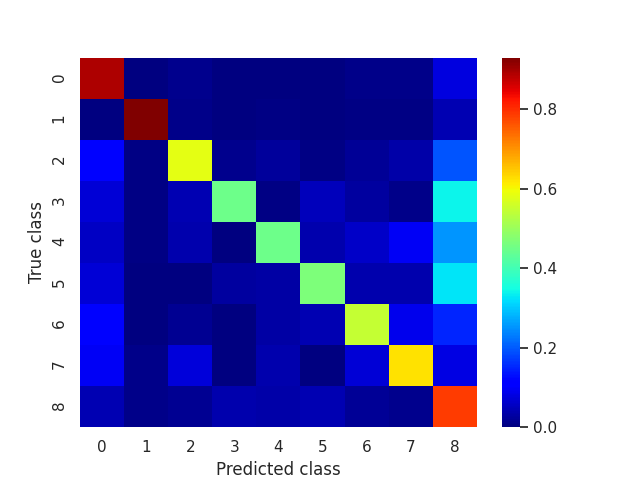}}
		\subfigure[\textbf{CoCa (ours)}]{\label{fig:sccf}
			\includegraphics[scale=0.33,trim=10 0 40 20,clip]{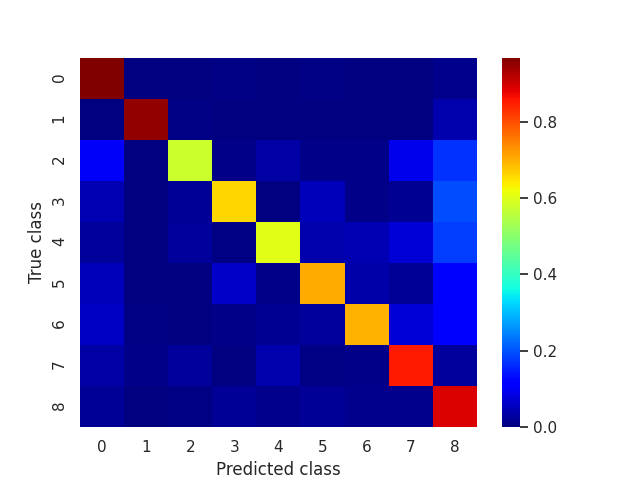}}
	\end{center}
	\caption{Confusion matrices of four different variations for test set of MNIST-360 dataset.  (a) Confusion matrix of JOINT. (b) Confusion matrix of SGD. (c) Confusion matrix of ER. (d) Confusion matrix of CoCa.}
	\label{fig:cfm}
\end{figure*}
\subsection{Ablation Study}
\textit{1) The impact of each component: } To evaluate the impact of different components in CoCa, we conduct ablation studies on sequential CIFAR-100 and MNIST-360 datasets, as shown in Table \ref{ablationstudy}. We take the Experience Replay Module as baseline, upon which the following components are considered: CKD applies collaborative distillation loss, PT adds  pretext tasks loss, SCL introduces supervised contrastive learning loss, CSS represents the combination of  pretext tasks loss and supervised contrastive learning loss. \par
As shown in Table \ref{ablationstudy}, each component contributes positively to the model except for the  PT on the MNIST-360 dataset. This is because the samples in the MNIST-360 dataset are rotated at increasing angles, which interferes with the prediction of the pretext tasks. However, its combination with SCL, i.e., CSS, performs better than each of them individually, which proves that the PT and SCL complement each other. Furthermore,  when all the components are combined, the best performance is achieved in all settings. This consistent improvement verifies our statement that the joint calibration between CKD and CSS is beneficial for alleviating the deviation, that is, the modules of relation calibration and feature calibration are mutually complementary. \par
\begin{table}[]
\caption{Ablation Studies on CoCa components on sequential CIFAR-100 and MNIST-360  datasets. 'CKD' indicates collaborative knowledge distillation, 'PT' indicates pretext tasks and 'SCL' indicates supervised contrastive learning} \label{ablationstudy}
\begin{center}	
\begin{tabular}{p{12 pt}p{12 pt}p{12 pt}p{16pt}p{16 pt}p{16 pt}p{16 pt}p{16 pt}p{16 pt}}
\toprule [1 pt]
\centering{CKD}&\centering{PT}&\centering{SCL}& \multicolumn{3}{c}{sequential CIFAR-100} &\multicolumn{3}{c}{MNIST-360} \\
 \multicolumn{3}{c}{$\mathcal{B}$ } & 200 & 500& 5120 & 200 & 500& 1000     \\ \hline
  &&  & 9.84  & 14.64 &44.79 &49.27 &65.04 &75.18  \\
 \centering{\checkmark} && & 20.26 & 30.27 & 51.55 & 56.71 & 71.07 &76.18 \\ 
  & \centering{\checkmark} &  & 10.79 &19.95 &47.38 &47.09 &62.24 &72.75 \\ 
 &&  \centering{\checkmark}& 11.33 & 16.44 & 51.11 & 59.37 & 66.90 & 75.73\\
 & \centering{\checkmark} & \centering{\checkmark} &12.25 &21.25 &52.31 &64.14 &70.83 &75.91 \\
 \centering{\checkmark} & \centering{\checkmark} & & 20.77 & 30.43 & 54.29 &53.59 &71.35 &77.49 \\ 
 \centering{\checkmark} && \centering{\checkmark} & 18.60 & 30.65 & 53.19 &66.46 &74.75 &81.28\\  
 \centering{\checkmark}& \centering{\checkmark}& \centering{\checkmark} & \textbf{21.20} & \textbf{32.88} & \textbf{58.38} & \textbf{67.02} & \textbf{76.49} & \textbf{81.82}\\ \toprule [1 pt]
\end{tabular}
\end{center}
\end{table}
\textit{2) The impact of the different numbers of tasks: } We further conduct experiments to explore the impact of the different numbers of tasks. We choose sequential CIFAR-100 dataset as example, which includes 10-spilt and 20-spilt settings, that is, dividing the CIFAR-100 dataset into 10 and 20 tasks, respectively. As shown in Table \ref{CIFAR100-task}, we observe that all competitors except for the upper bound JOINT are sensitive to the number of tasks. As the number of tasks increases from 10 to 20, the performance drops a lot. For example, when buffer sizes are 200, 500 and 5120, the declines are 4.86\%, 5.55\% and 2.86\% in CoCa, respectively. This is due to that the increasing number of the tasks leads to the decreasing number of the classes for each task.  However, in the case of either a 10-split or 20-split settings, our approach is superior to the other competitors, demonstrating that the effectiveness of CoCa framework. \par
\begin{table}[]
\caption{ Ablation Studies on the different numbers of tasks on sequential CIFAR-100 dataset} \label{CIFAR100-task}
\begin{center}
\begin{tabular}{ccccccc}
\toprule [1 pt]
Method & \multicolumn{3}{c}{10 Tasks}   & \multicolumn{3}{c}{20 Tasks}  \\ \hline
JOINT & \multicolumn{3}{c}{69.55} & \multicolumn{3}{c}{69.55} \\
SGD & \multicolumn{3}{c}{8.54}  & \multicolumn{3}{c}{4.33}  \\ \hline
$\mathcal{B}$  & 200 & 500 & 5120 & 200 & 500 & 5120\\ \hline
ER \cite{ser} &13.81  &21.81 & 49.83 & 9.84   & 14.64  & 44.79\\
DER \cite{NEURIPS2020_DER} &23.25 &36.20 & 56.20& 18.66 & 28.70  & 51.20\\
\textbf{CoCa (ours)}   & \textbf{26.06} & \textbf{38.43} & \textbf{61.25} & \textbf{21.20} & \textbf{32.88} & \textbf{58.38} \\ \toprule [1 pt]
\end{tabular}
\end{center}
\end{table}
\textit{3) The impact of parameters $\omega$ and $\gamma$: } Hyper-parameters $\omega$ and $\gamma$ are important  in calibrating the relation deviation in Eq.\ref{eq11}. We also select sequential CIFAR-100 dataset as example, whose results are shown in Fig.\ref{fig:sensitivity}.  As observed from Fig. \ref{fig:omega},  the different proportions $\omega$ of label propagation have little impact on Relation Calibration Module within the interval of $(0.1, 0.9)$. However, when $\omega$ reaches 1, there is a significant decline.  This is because the ensemble output $o^*$ is equal to the reserved output $\hat{o}$, which fails to explore  collaborative distillation. Moreover, as shown in Fig. \ref{fig:gamma}, $\gamma$ is a sensitive hyper-parameter on balancing the relationship among all classes in collaborative distillation. As we can see, even if $\gamma$ is 0.1, the accuracy improves more than 2\% against that of 0. This explains that it is necessary to alleviate the relation deviation in knowledge distillation. However, as $\gamma$  increases, the ensemble output $o^*$ contains less information from the old model, resulting in a dramatic decline in model performance. \par
\subsection{Visualization  Analysis}
\textit{1) t-SNE Results: } To further verify the effectiveness of our CoCa framework on calibrating the feature deviation, we visualize the features for the test set of MNIST-360 dataset, which are shown in Fig. \ref{fig:tsne}. We observe that the t-SNE of the baseline method ER is better than that of SGD. However, as evident in Fig. \ref{fig:ert}, classes are not distinguished well and the class boundaries are also not precise and compact. The Relation Calibration Module is formed by adding the collaborative distillation loss to the baseline, whose t-SNE is shown in Fig. \ref{fig:rct}. As we can see, compared with ER, Relation Calibration Module reduces the interference of features among different classes through the normalized similarity matrix, which further helps to aggregate samples of the same class. Figure \ref{fig:fct} shows the t-SNE corresponding to Feature Calibration Module, from which we observe that each class is well distinguished and the boundaries of most classes are clear. On the one hand, the complete features are obtained through pretext tasks, leading to the class representations are well spread out. On the other hand, the supervised contrastive learning enables the inter-class distance larger and the intra-class distance smaller, which makes the boundaries are clearer. Finally, our method takes advantage of the complementary modules that lead to more compact clusters and discriminative class boundaries. It is worth noting that the t-SNE of our method is comparable to that of JOINT, which also proves the effectiveness of our CoCa framework in alleviating the feature deviation.\par
\textit{2) Visualization of confusion matrices: } Figure \ref{fig:cfm} provides the visualization of confusion matrices on the test set of MNIST-360 dataset to give an insight into the effectiveness of our CoCa framework. Among them, diagonal entries represent the accuracy of each class. As shown in  Fig. \ref{fig:sgdcf},  SGD is obviously biased towards the last task (8,0). Interestingly, it obtains an easily distinguishable feature for the number 1 (see Fig. \ref{fig:sgdt}), but its accuracy is still zero. This proves that the catastrophic forgetting is the result of feature extractor and classifier. By comparing Fig. \ref{fig:ercf}  and Fig. \ref{fig:sccf}, we observe that the CoCa has few deviations towards the last task and achieves superior performance among all classes.

\section{Conclusion}
In this paper, we have proposed the CoCa framework to alleviate the relation and feature deviations in GCL by collaborative distillation and self-supervision. Specifically, the collaborative distillation mitigates the relation deviation by exploring ensemble dark knowledge in knowledge distillation  to balance the relationship among classes. The collaborative self-supervision is composed by pretext tasks and supervised contrastive learning,  which aims at learning complete and discriminative features to alleviate the feature deviation.  Extensive experiments have demonstrated that our proposed CoCa framework outperforms the state-of-the-art ones. In future, we consider to leverage online distillation approaches and explore how to select positive samples in contrastive learning. \\

\ifCLASSOPTIONcaptionsoff
  \newpage
\fi



%

\bibliographystyle{IEEEtran}
\bibliography{CoCa}

%




\end{document}